\documentclass[12pt]{article}
\usepackage[final]{nips_2017}

\usepackage[utf8]{inputenc} 
\usepackage[T1]{fontenc}    
\usepackage{hyperref}       
\usepackage{url}            
\usepackage{booktabs}       
\usepackage{amsfonts}       
\usepackage{nicefrac}       
\usepackage{microtype}      
\usepackage{float}

\usepackage{amsmath} 
\usepackage{amssymb}
\usepackage{amsthm}
\usepackage{mathrsfs}
\usepackage{enumerate}
\usepackage{pgf}
\usepackage{tikz}
\usetikzlibrary{arrows,automata}
\usepackage{framed}
\usepackage[,slantedGreek]{mathptmx}
\usepackage{dsfont}

\usepackage{algorithm, algorithmic}
\floatname{algorithm}{\texttt{AGB} algorithm}

\usepackage{natbib}
\bibpunct{(}{)}{;}{a}{,}{,}
\usepackage{hyperref}
\hypersetup{
colorlinks = true,
urlcolor = blue, 
linkcolor = blue,
citecolor = blue,
}

\title{Accelerated Gradient Boosting}

\author{
  G.~Biau\\
  Sorbonne Universit\'e, CNRS, LPSM\\
  Paris, France \\
  \texttt{gerard.biau@upmc.fr} \\
 \And
  B.~Cadre\\
  Univ Rennes, CNRS, IRMAR \\
  Rennes, France \\
   \texttt{benoit.cadre@ens-rennes.fr} \\
\And
 L.~Rouvi\`ere\\
  Univ Rennes, CNRS, IRMAR \\
  Rennes, France \\
   \texttt{laurent.rouviere@univ-rennes2.fr} \\
}

\begin{document}

\maketitle

\begin{abstract}
Gradient tree boosting is a prediction algorithm that
sequentially produces a model in the form of linear combinations of
decision trees, by solving an infinite-dimensional optimization problem. 
We combine gradient boosting and
Nesterov's accelerated descent to design a new algorithm,
which we call \verb+AGB+ (for Accelerated Gradient Boosting). Substantial numerical evidence 
is provided on both synthetic and real-life data sets to assess the excellent performance 
of the method in a large variety of prediction problems.
It is empirically shown that \verb+AGB+ is much less sensitive to the shrinkage parameter and
outputs predictors that are considerably more sparse in the number 
of trees, while retaining the exceptional performance of gradient boosting. 
\end{abstract}

\section{Introduction}
Gradient boosting \citep{FrHaTi00,Fr01,Fr02} is a learning procedure that combines the outputs of many simple predictors in order to produce a powerful committee with performances improved over the single members. The approach is typically used with decision trees of a fixed size as base learners, and, in this context, is called gradient tree boosting. This machine learning method is widely recognized for providing state-of-the-art results on several challenging data sets, as pointed out for example in the introduction of \citet{ChGu16}. To get to the point, boosted decision trees are generally regarded as one of the best off-the-shell prediction algorithms we have today, with performance at the level of the Lasso \citep{Ti96} and random forests \citep{Brforests01}, to name only two competitors. 

Gradient boosting originates in Freund and Schapire's work \citep{Sc90,Fr95,FrSc96a,FrSc97} on weighted iterative classification. It was complemented by several analyses by \citet{Br97,Br98,Br99,Br00,Br04}, who made the fundamental observation that Freund and Schapire's AdaBoost is in fact a gradient-descent-type algorithm in a function space, thus identifying boosting at the frontier of numerical optimization and statistical estimation. Explicit regression and classification boosting algorithms were subsequently developed by \citet{Fr01,Fr02}, who coined the name ``gradient boosting'' and paid a special attention to the case where the individual components are decision trees. Overall, this functional view of boosting has led to the development of boosting algorithms in many areas of machine learning and statistics beyond regression and classification \citep[e.g.,][]{BlLuVa03,BuYu03,LuVa04,ZhYu05,BiRiZa06,BuHo07}.

In a different direction, the pressing demand of the machine learning community to build accurate prediction mechanisms from massive amounts of high dimensional data has greatly promoted the theory and practice of accelerated first-order schemes. In this respect, one of the most effective approaches among first-order optimization techniques is the so-called Nesterov's accelerated gradient descent \citep{Ne83}. In a nutshell, if we are interested in minimizing some smooth convex function $f(x)$ over $\mathds R^d$, then Nesterov's descent may take the following form \citep{BeTe09}: starting with $x_0=y_0$, inductively define
\begin{equation} 
\label{nesterov} 
\begin{array}{lll}
x_{t+1} & = &y_{t}-w \nabla f(y_t)\\
y_{t+1} & =  & (1-\gamma_t)x_{t+1}+\gamma_t x_t,
\end{array}
\end{equation}
where $w$ is the step size, 
$$\lambda_0=0, \quad \lambda_t=\frac{1+\sqrt{1+4 \lambda_{t-1}^2}}{2}, \quad \mbox{and} \quad \gamma_t=\frac{1-\lambda_t}{\lambda_{t+1}}.$$
In other words, Nesterov's descent performs a simple step of gradient to go from $y_t$ to $x_{t+1}$, and then it slides it a little bit further than $x_{t+1}$ in the direction given by the previous point $x_t$. As acknowledged by \citet{Bu13}, the intuition behind the algorithm is quite difficult to grasp. Nonetheless, Nesterov's accelerated gradient descent is an optimal method for smooth convex optimization: the sequence $(x_t)_t$ recovers the minimum of $f$ at a rate of order $1/t^2$, in contrast to vanilla gradient descent methods, which have the same computational complexity but can only achieve a rate in ${\rm O}(1/t)$. Since the introduction of Nesterov's scheme, there has been much work on first-order accelerated methods (\citealp[see, e.g.,][for theoretical developments]{Ne04,Ne05,Ne13,SuBoCa16}, and \citealp{Ts08}, for a unified analysis of these ideas). Notable applications can be found in sparse linear regression \citep{BeTe09}, compressed sensing \citep{BeBoCa11}, distributed gradient descent \citep{QuLi17}, and deep and recurrent neural networks \citep{SuMaDaHi13}.

\begin{figure}[h]
  \centering
  \includegraphics[width=13cm,height=13cm]{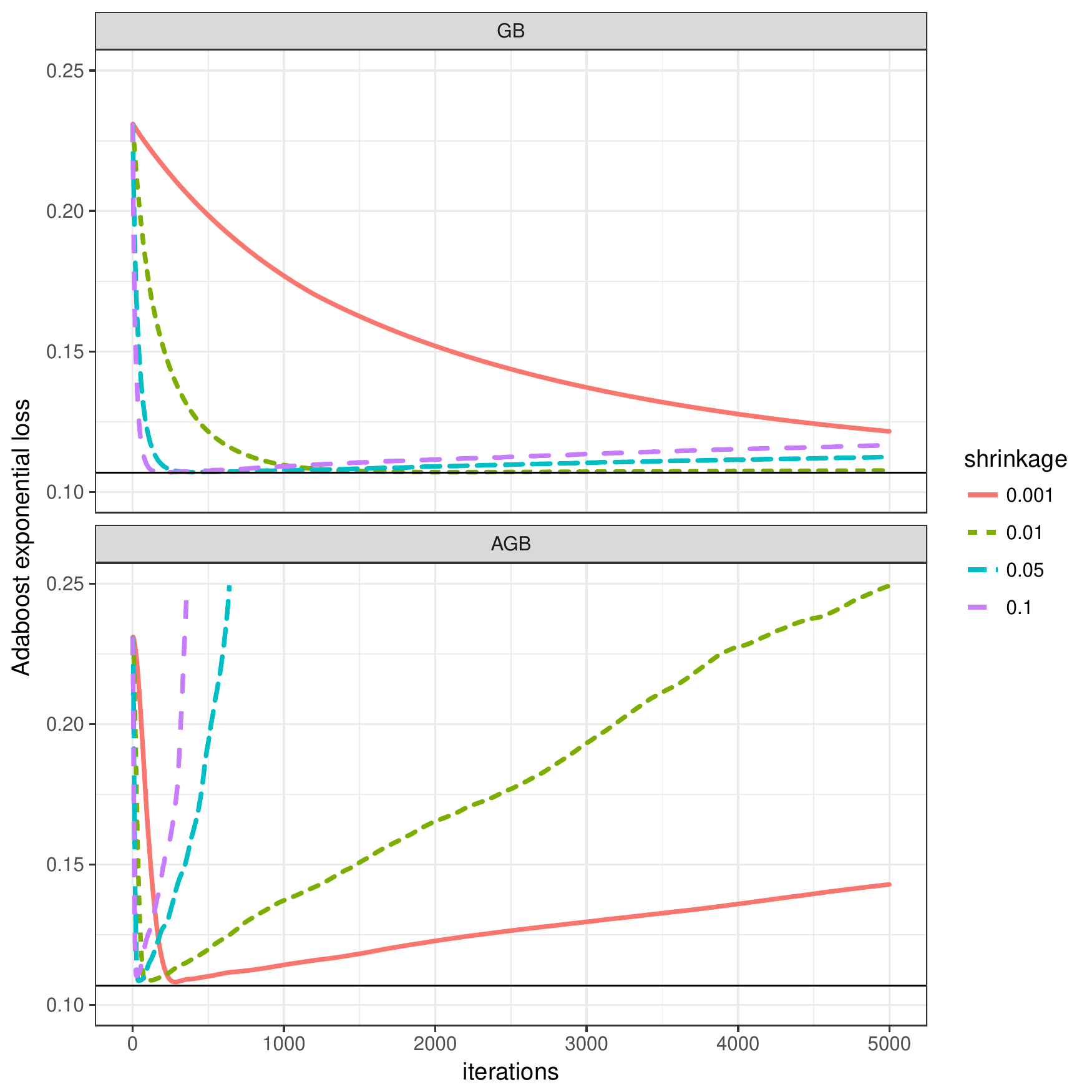}
  \caption{Adaboost exponential loss (estimated on a test data set) by number of iterations for standard gradient boosting (top) and \texttt{AGB} (bottom). The data are generated according to \texttt{Model 5} with $n=5\ 000$ observations (see page \pageref{def:model5}).}
  \label{fig:err_plusieurs_shrink}
\end{figure}

In this article, we present \verb+AGB+ (for \verb+A+ccelerated \verb+G+radient \verb+B+oosting), a new tree boosting algorithm that incorporates Nesterov's mechanism (\ref{nesterov}) into Friedman's original procedure \citep{Fr01}.  Substantial numerical evidence is provided on both synthetic and real-life data sets to assess the excellent performance of our method in a large variety of prediction problems. The striking feature of \verb+AGB+ is that it enjoys the merits of both approaches: 
\begin{enumerate}[$(i)$]
\item Its predictive performance is comparable to that of standard gradient tree boosting;
\item It takes advantage of the accelerated descent to output models which are remarkably 
much more sparse in their number of components. 
\end{enumerate}
Item $(ii)$ is of course a decisive advantage for large-scale learning, when time and storage issues matter. To make the concept clear, we show in Figure \ref{fig:err_plusieurs_shrink} typical test error results by number of iterations and shrinkage (step size), both for the standard (top) and the accelerated (bottom) algorithms. As is often the case with gradient boosting, smaller values of the shrinkage parameter require a larger number of trees for the optimal model, when the test error is at its minimum. However, if both approaches yield similar results in terms of prediction, we see that the optimal number of iterations is at least one order of magnitude smaller for \verb+AGB+.

The paper is organized as follows. In Section \ref{(A)GB}, we briefly recall the mathematical/statistical context of gradient boosting, and present the principle of the \verb+AGB+ algorithm. Section \ref{NS} is devoted to analyzing the results of a battery of experiments on synthetic and real-life data sets. We offer an extensive comparison between the performance of Friedman's gradient tree boosting and \verb+AGB+, with a special emphasis put on the influence of the learning rate on the size of the optimal models. The code used for the simulations and the figures is available at \url{https://github.com/lrouviere/AGB}. 
\section{(Accelerated) gradient boosting}
\label{(A)GB}
\subsection{Gradient boosting at a glance}
Let $\mathscr D_n=\{(X_1,Y_1), \hdots, (X_n,Y_n)\}$ be a sample of i.i.d.~observations, all distributed as an independent generic pair $(X,Y)$ taking values in $\mathds R^d \times \mathscr Y$. Throughout, $\mathscr Y\subset \mathds R$ is either a finite set of labels (for classification) or a subset of $\mathds R$ (for regression). The learning task is to construct a predictor $F:\mathds R^d\to \mathds R$ that assigns a response to each possible value of the independent random observation $X$.  In the context of gradient boosting, this general problem is addressed by considering a class $\mathscr F$ of elementary functions $f:\mathds R^d\to \mathds R$ (called the weak or base learners), and by minimizing some empirical risk functional
\begin{equation}
\label{CNF}
C_n(F)=\frac{1}{n}\sum_{i=1}^n\psi(F(X_i),Y_i)
\end{equation}
over the linear combinations of functions in $\mathscr F$. Thus, we are looking for an additive solution of the form  $F_n=\sum_{j=0}^J \alpha_jf_j$, where $(\alpha_0, \hdots, \alpha_J) \in \mathds R^{J+1}$ and each component $f_j$ is picked in the base class $\mathscr F$.

The function $\psi:\mathds R \times \mathscr Y \to \mathds R_+$ is called the loss. It is assumed to be convex and differentiable in its first argument, and it measures the cost incurred by predicting $F(X_i)$ when the answer is $Y_i$. For example, in the least squares regression problem, $\psi(x,y)=(y-x)^2$, and
$$C_n(F)=\frac{1}{n}\sum_{i=1}^n(Y_i-F(X_i))^2.$$
In the $\pm1$-classification problem, the final classification rule is $+1$ if $F(x)>0$ and $-1$ otherwise. In this context, two classical losses are $\psi(x,y)=e^{-yx}$ (Adaboost exponential loss) and $\psi(x,y)=\ln_2(1+e^{-yx})$ (logit loss).

\begin{figure}[H]
  \centering
  \includegraphics[width=15cm,height=6cm]{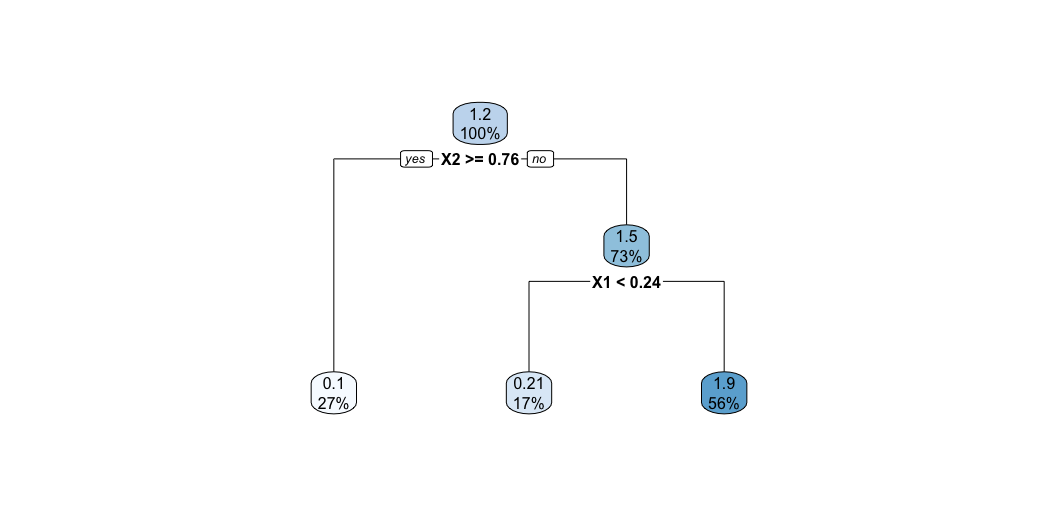}
  \caption{A regression tree in dimension $d=2$ with $k=3$ leaves.}
  \label{fig:reg_tree}
\end{figure}

In the present document, we take for $\mathscr F$ the collection of all binary decision trees in $\mathds R^d$ using axis parallel cuts with $k$ (small) terminal nodes (or leaves). Thus, each $f \in \mathscr F$ takes the form $f=\sum_{j=1}^{k} \beta_j \mathds 1_{A_j}$, where 
$(\beta_1, \hdots, \beta_k) \in \mathds R^k$ and $\{A_1, \hdots, A_{k}\}$ is a tree-structured partition of $\mathds R^d$ \citep[][Chapter 20]{DeGyLu96}. An example of regression tree fitted with the \verb+R+ package \verb+rpart.plot+ with $k=3$ leaves in dimension $d=2$ is shown in Figure \ref{fig:reg_tree}.

Let us get back to the minimization problem (\ref{CNF}) and denote by ${\mbox{lin}(\mathscr F)}$ the set of all linear combinations of functions in $\mathscr F$, our basic collection of trees. So, each $F \in {\mbox{lin}(\mathscr F)}$ is an additive association of trees, of the form $F=\sum_{j=0}^J \alpha_j f_j$. Finding the infimum of the functional $C_n$ over ${\mbox{lin}(\mathscr F)}$ is a challenging infinite-dimensi\-o\-nal optimization problem, which requires an algorithm. This is where gradient boosting comes into play by sequentially constructing a linear combination of trees, adding one new component at each step. This algorithm rests upon a sort of functional gradient descent, which we briefly describe in the next paragraph. We do not go to much into the mathematical details, and refer to \citet{MaBaBaFr99,MaBaBaFr00} and \citet{BiCa17} for a thorough analysis of the mathematical forces in action. 
 
Suppose that we have at step $t$ a function $F_t \in \mbox{lin}(\mathscr F)$ and wish to find a new $f_{t+1}\in \mathscr F$ to add to $F_t$ so that the risk $C_n(F_t+wf_{t+1})$ decreases at most, for some small value of $w$. Viewed in function space terms, we are looking for the direction $f_{t+1} \in \mathscr F$ such that $C_n(F_t+wf_{t+1})$ most rapidly decreases. Observe that, for all $F \in \mbox{lin}(\mathscr F)$, $\nabla C_n (F)(X_i)=\partial_x\psi(F(X_i),Y_i)$, where the symbol $\partial_x$ means partial derivative with respect to the first component. Then the knee-jerk reaction is to take $f_{t+1}(\cdot)=-\nabla C_n (F_t)(\cdot)$, the opposite of the gradient of $C_n$ at $F_t$ (this is a function over $\mathds R^d$), and do something like
$$F_{t+1}=F_t-w\nabla C_n (F_t).$$
However, since we are restricted to pick our new function in $\mathscr F$, this will in general not be a possible choice. The stratagem is to choose the new $f_{t+1}$ by a least squares approximation of the function $-\nabla C_n (F_t)(\cdot)$, i.e., to take
$$f_{t+1}\in {\arg \min}_{f \in \mathscr F} \frac{1}{n}\sum_{i=1}^n (-\nabla C_n (F_t)(X_i)-f(X_i))^2.$$
For example, when $\psi(x,y)=(y-x)^2/2$, then $-\nabla C_n(F_t)(X_i)=Y_i-F_t(X_i)$, and the algorithm simply fits $f_{t+1}$ to the residuals $Y_i-F_t(X_i)$ at step $t$. This is the general principle of Friedman's gradient boosting  \citep{Fr01}, which after $T$ iterations outputs an additive expansion of the form $F_T=\sum_{t=0}^T\alpha_tf_{t}$. The operational algorithm includes several regularization techniques to reduce the eventual overfitting. Some of these features are incorporated in our accelerated version, which we now describe.
\subsection{The AGB algorithm}
The pseudo-code of \verb+AGB+ is presented in the table below. 
\begin{algorithm}[H]
\label{algorithm1}
\caption{}
\begin{algorithmic}[1]
\medskip

\STATE {\bf Require} $T\geq 1$ (number of iterations), $k\geq 1$ (number of terminal nodes in the trees), $0<\nu<1$ (shrinkage parameter).
\medskip

\STATE {\bf Initialize} $F_0=G_0={\arg \min}_{z} \sum_{i=1}^n \psi(z,Y_i)$, $\lambda_0=0$, $\gamma_0=1$.
\medskip

\FOR{$t=0$ to $(T-1)$} 
\medskip

\STATE For $i=1, \hdots, n$, {\bf compute} the negative gradient  instances
$$
Z_{i,t+1}=-\nabla C_n (G_t)(X_i).
$$
\STATE {\bf Fit} a regression tree to the pairs $(X_i,Z_{i,t+1})$, giving terminal nodes $R_{j,t+1}$, $1 \leq j \leq k$.
\medskip

\STATE For $j=1, \hdots, k$, {\bf compute}
$$w_{j,t+1}\in{\arg \min}_{w>0} \sum_{X_i \in R_{j,t+1}} \psi(G_{t}(X_i)+w,Y_i).$$
\STATE {\bf Update}
\medskip
\begin{enumerate}[$(a)$]
\item $F_{t+1}=G_{t}+\nu\sum_{j=1}^{k} w_{j,t+1} \mathds 1_{R_{j,t+1}}$.
\smallskip
\item $G_{t+1}=(1-\gamma_{t})F_{t+1}+\gamma_t F_t$.
\smallskip
\item $\lambda_{t}=\frac{1+\sqrt{1+4 \lambda_{t-1}^2}}{2}$, $\lambda_{t+1}=\frac{1+\sqrt{1+4 \lambda_{t}^2}}{2}$.
\smallskip
\item $\gamma_{t}=\frac{1-\lambda_{t}}{\lambda_{t+1}}$.
\end{enumerate}
\medskip
\ENDFOR
\medskip

\STATE {\bf Output} $F_{T}$.
\end{algorithmic}
\end{algorithm}
We see that the algorithm has two inner functional components, $(F_t)_t$ and $(G_t)_t$, which correspond respectively to the vectorial sequences $(x_t)_t$ and $(y_t)_t$ of Nesterov's acceleration scheme (\ref{nesterov}). Observe that the sequence $(G_t)_t$ is internal to the procedure while the linear combination output by the algorithm after $T$ iterations is $F_T$. Line 2 initializes to the optimal constant model. As in Friedman's original approach, the algorithm selects at each iteration, by least-squares fitting, a particular tree that is in most agreement with the descent direction (the ``gradient''), and then performs an update of $G_t$. The essential difference is the presence of the companion function sequence $(G_t)_t$, which slides the iterates $(F_t)_t$ according to the recursive parameters $\lambda_t$ and $\gamma_t$ (lines $7$ $(b)$-$(d)$).
 
Let $f_{t+1}=\sum_{j=1}^k{\beta_{j,t+1}}\mathds 1_{R_{j,t+1}}$ be the approximate-gradient tree output at line 6 of the algorithm. The next logical step is to perform a line search to find the step size and update the model accordingly, as follows:
$$w_{t+1}\in {\arg \min}_{w>0}\,\sum_{i=1}^n\psi(G_{t}(X_i)+w f_{t+1}(X_i), Y_i), \quad F_{t+1}=G_{t}+w_{t+1} f_{t+1}.$$
However, following Friedman's gradient tree boosting \citep{Fr01}, a separate optimal value $w_{j,t+1}$ is chosen for each of the tree's regions, instead of a single $w_{t+1}$ for the whole tree. The coefficients $\beta_{j,t+1}$ from the tree-fitting procedure can be then simply discarded, and the model update rule at epoch $t$ becomes, for each $j=1, \hdots, k$,
$$w_{j,{t+1}} \in {\arg \min}_{w>0} \sum_{X_i \in R_{j,t+1}} \psi(G_{t}(X_i)+w,Y_i), \quad F_{t+1}=G_{t}+\nu\sum_{j=1}^k w_{j,t+1}\mathds 1_{R_{j,t+1}}$$
(lines 6 and 7 $(a)$). We also note that the contribution of the approximate gradient is scaled by a factor $0<\nu<1$ when it is added to the current approximation. The parameter $\nu$ can be regarded as controlling the learning rate of the boosting procedure. Smaller values of $\nu$ (more shrinkage) usually lead to larger values of $T$ for the same training risk. Therefore, in order to reduce the number of trees composing the boosting estimate, large values for $\nu$ are required. However, too large values of $\nu$ may break the gradient descent dynamic, as shown for example in \citet[][Lemma 3.2]{BiCa17}. All in all, both $\nu$ and $T$ control prediction risk on the training data and these parameters do not operate independently. This tradeoff issue is thoroughly explored in the next section.
\section{Numerical studies}
\label{NS}
This section is devoted to illustrating the potential of our \verb+AGB+ algorithm and to highlighting the benefits of Nesterov's acceleration scheme in the boosting process. Synthetic models and real-life data are considered, and an exhaustive comparison with standard gradient tree boosting is performed. For the implementation of Friedman's boosting, we used the \verb+R+ package \verb+gbm+, a description of which can be found in \citet{Ri07}. These two boosting algorithms are compared in the last subsection with the Lasso \citep{Ti96} and random forests \citep{Brforests01} methods, respectively implemented with the packages \verb+glmnet+ and \verb+randomForest+.
\subsection{Description of the data sets}
The algorithms were benchmarked on both simulated and real-life data sets. For each of the simulated models, we consider two designs for $X=(X_1,\hdots,X_d)$: Uniform over $(-1,1)^d$ ("Uncorrelated design") and Gaussian with mean $0$ and $d\times d$ covariance matrix $\Sigma$ such that $\Sigma_{ij}=2^{-|i-j|}$ ("Correlated design'').
The five following models cover a wide spectrum of regression and classification problems. Models 1-3 and 5 come from \cite{BiFiGuMa16}. Model 4 is a slight variation of a benchmark model in \cite{HaTiFr09}. Models 1-3 are regression problems, while Model 4 and 5 are $\pm 1$-classification tasks. Models 2-4 are additive, while Models 1 and 5 include some interactions. Model 3 can be seen as a sparse high-dimensional problem. We denote by $Z_{\mu,\sigma^2}$ a Gaussian random variable with mean $\mu$ and variance $\sigma^2$.

\paragraph{Model 1.} $n=1\,000$, $d=100$, $Y=X_1X_2+X_3^2-X_4X_7+X_8X_{10}-X_6^2+Z_{0,0.5}$.
\paragraph{Model 2.} $n=800$, $d=100$, $Y=-\sin(2X_1)+X_2^2+X_3-\exp(-X_4)+Z_{0,0.5}$.
\paragraph{Model 3.} $n=1\,000$, $d=500$, $Y=X_1+3X_3^2-2\exp(-X_5)+X_6$.
\paragraph{Model 4.} $n=2\,000$, $d=30$,
$$Y=\left\{
  \begin{array}{ll}
2\ \mathds{1}_{\sum_{j=1}^{10}X_j^2> 3.5}-1  & \text{for uncorrelated design} \\
2\ \mathds{1}_{\sum_{j=1}^{10}X_j^2> 9.34}-1  & \text{for correlated design}.
  \end{array}\right.
$$
\paragraph{Model 5.} \label{def:model5}$n=1\,500$, $d=50$, $Y=2\ \mathds{1}_{X_1+X_4^3+X_9+\sin(X_{12}X_{18})+ Z_{0,0.1}>0.38}-1$.

We also considered the following real-life data sets from the \texttt{UCI Machine Learning repository}: Adult, Internet Advertisements, Communities and Crime, Spam, and Wine. Their main characteristics are summarized in Table \ref{tab:descdonreelles} (a more complete description is available at the address \url{https://archive.ics.uci.edu/ml/datasets.html}).
\begin{table}[h]
  \centering
  \begin{tabular}{|c||c|c|c|}
\hline
\hline
{\bf Data set}  & $n$ & $d$ & {\bf Output} $Y$ \\
\hline\hline
Adult & 30\,162 & 14 & binary \\
Advert. & 2\,359 &  1\,431 & binary \\
Crime & 1\,993 & 102 & continuous \\
Spam & 4\,601 & 57 & binary \\
Wine & 1\,559 & 11 & continuous \\
\hline
\hline
  \end{tabular}
  \medskip
  \caption{Main characteristics of the five real-life data sets used in the experiments.}
  \label{tab:descdonreelles}
\end{table}

For each data set, simulated or real, the sample is divided into a training set (50\%) $\mathscr D_{{\rm train}}$ to fit the method; a validation set (25\%) $\mathscr D_{{\rm val}}$ to select the hyperparameters of the algorithms; and a test set (25\%) $\mathscr D_{{\rm test}}$ on which the predictive performance is evaluated.
We considered two loss functions for both standard boosting and \verb+AGB+: the least squares loss $\psi(x,y)=(y-x)^2$ for regression and the Adaboost loss $\psi(x,y)=e^{-yx}$ for $\pm 1$-classification. We also tested the logit loss function $\psi(x,y)=\ln_2(1+e^{-yx})$. Since the results are similar to the Adaboost loss  they are not reported. 

In the boosting algorithms, the validation set is used to select the number of components of the model, i.e., the number of iterations performed by the algorithm. Thus, denoting by $F_T$ the boosting predictor after $T$ iterations fitted on $\mathscr D_{{\rm train}}$, we select the $T^\star$ that minimizes
\begin{equation}
  \label{eq:sel_iter_test}
  \frac{1}{\sharp \mathscr D_{{\rm val}} }\sum_{i\in\mathscr D_{{\rm val}}}\psi(F_T(X_i),Y_i).
\end{equation}
For both standard gradient tree boosting and \verb+AGB+, we fit regression trees with two terminal nodes. We considered five fixed values for the shrinkage parameter $\nu$ ($1\mathrm{e}-05$, 0.001, 0.01, 0.1, and 0.5), and fixed an arbitrary (large) limit of $T=10\,000$ iterations for the standard boosting and $T=2\,500$ for \verb+AGB+. All results are averaged over 100 replications for simulated examples, and over 20 independent permutations of the sample for the real-life data.
\subsection{Gradient boosting vs accelerated gradient boosting}
\begin{figure}[h]
  \centering
 \includegraphics[width=13cm,height=10cm]{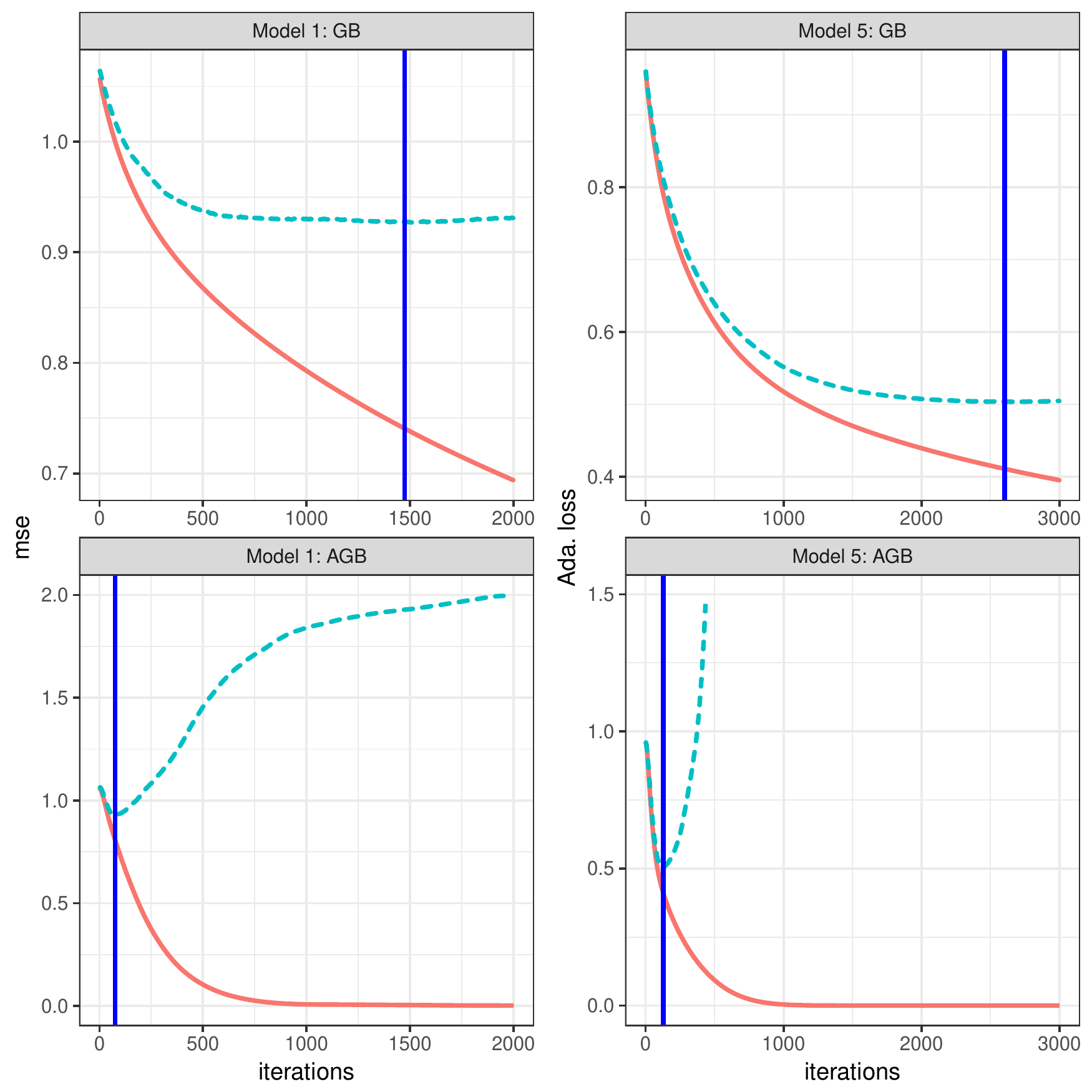}
  \caption{Training (solid lines) and validation (dashed lines) errors for Model 1 and Model 5. Shrinkage parameter $\nu$ is fixed to 0.01.}
  \label{fig:sel_nb_iter}
\end{figure}

In this subsection, we compare the standard gradient tree boosting and \verb+AGB+ algorithms in terms of minimization of the empirical risk \eqref{CNF} and selected number of components $T^{\star}$. Figure \ref{fig:sel_nb_iter} shows the training and validation errors for Friedman's boosting and \verb+AGB+ (bottom), as a function of the number of iterations.  

As it is generally the case for gradient boosting \citep[e.g.,][]{Ri07}, the validation error decreases until predictive performance is at its best and then starts increasing again. The vertical blue line shows the optimal number of iterations $T^{\star}$, selected by minimizing \eqref{eq:sel_iter_test}. We see that the validation rates at the optimal $T^{\star}$ are comparable for \verb+AGB+ and the original algorithm. However, \verb+AGB+ outperforms gradient boosting in terms of number of components of the output model, which is much smaller for \verb+AGB+. This is a direct consequence of Nesterov's acceleration scheme.

\begin{figure}[!h]
  \centering
  \includegraphics[width=15cm,height=20cm]{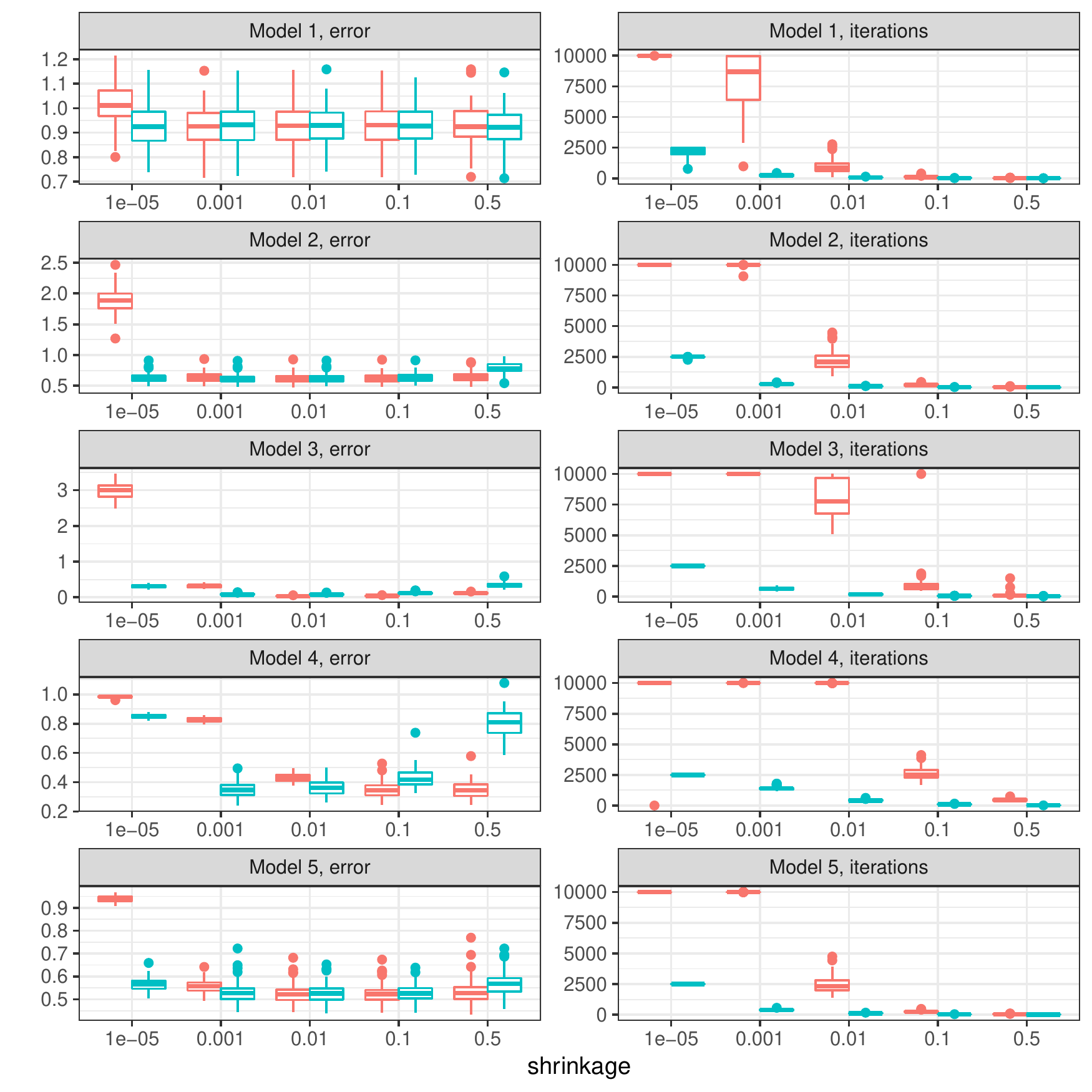}
  \caption{Boxplots of the test error \eqref{eq:dev_valid} (left) and selected numbers of iterations (right), as a function of the shrinkage parameter $\nu$ for standard gradient boosting (red, left) and \texttt{AGB} (blue, right). Results are presented for simulated models with uncorrelated design.}
  \label{fig:box_dev_reg1}
\end{figure}

\begin{figure}[!h]
  \centering
  \includegraphics[width=15cm,height=20cm]{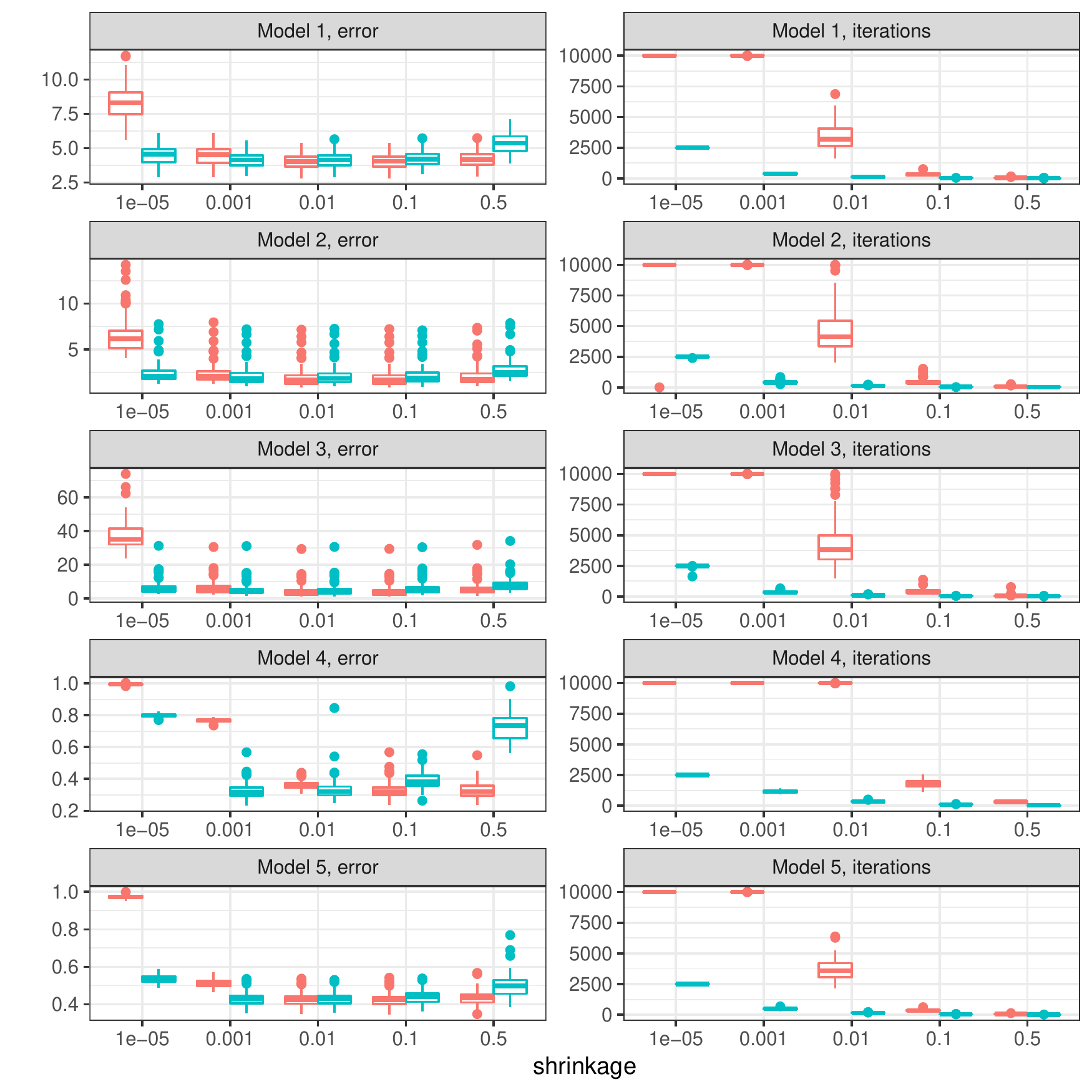}
  \caption{Boxplots of the test error \eqref{eq:dev_valid} (left) and number of selected iterations (right) as a function of the shrinkage parameter $\nu$, for standard gradient boosting (red, left) and \texttt{AGB} (blue, right). Results are presented for simulated models with correlated design.}
  \label{fig:box_dev_reg2}
\end{figure}

\begin{figure}[!h]
  \centering
  \includegraphics[width=15cm,height=20cm]{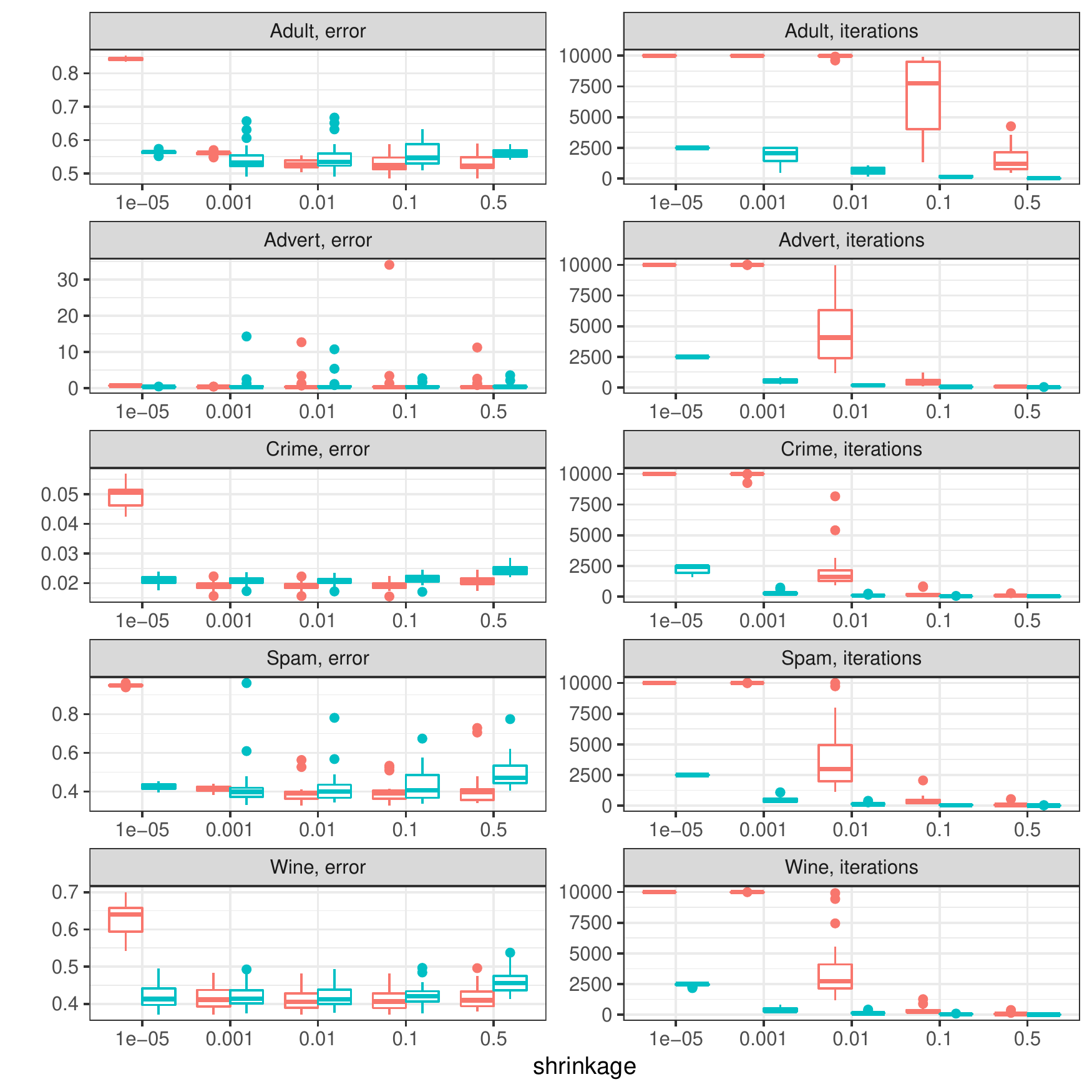}
  \caption{Boxplots of the test error \eqref{eq:dev_valid} (left) and number of selected iterations (right) as a function of the shrinkage parameter $\nu$, for standard gradient boosting (red, left) and \texttt{AGB} (blue, right). Results are presented for real-life data sets.}
  \label{fig:box_dev_reg3}
\end{figure}

This remarkable behavior is confirmed by Figures \ref{fig:box_dev_reg1}, \ref{fig:box_dev_reg2}, and \ref{fig:box_dev_reg3}, where we plotted the relationship between predictive performance, the number of iterations, and the shrinkage parameter. On the left side of each figure, we show the boxplots of the test errors of the selected predictors $F_{T^\star}$, i.e.,
\begin{equation}
  \label{eq:dev_valid}
\frac{1}{\sharp \mathscr D_{{\rm test}}}\sum_{i\in\mathscr D_{{\rm test}}}\psi(F_{T^\star}(X_i),Y_i),
\end{equation}
as a function of the shrinkage parameter $\nu$. The right sides depict the boxplots of the optimal number of components $T^\star$.

These three figures convey several messages. First of all, we notice that the predictive performances of the two methods are close to each other, independently of the data sets (simulated or real). Moreover, in line with the comments of \citet[][Chapter 10]{HaTiFr09}, smaller values of the shrinkage parameter $\nu$ favor better test error. Indeed, for all examples we observe that the best test errors are achieved for $\nu$ smaller than 0.1. However, for such values of $\nu$, it seems difficult for standard boosting to reach the optimal $T^{\star}$ in a reasonable number of iterations, and 10 000 iterations are generally not sufficient as soon as $\nu$ is less than 0.01. The accelerated algorithm allows to circumvent this problem since, for each value of $\nu$, the optimal model is achieved after a number of iterations considerably smaller than with standard boosting. Besides, \verb+AGB+ is much less sensitive to the choice of $\nu$. These two features are clear advantages since, in practice, one has no or few a priori information on the reasonable value of $\nu$, and the usual strategy is to try several (often, small) values of the shrinkage parameter until the validation error is the lowest.
Of course, this benefit is striking when we are faced with large-scale data, i.e., when iterations have a computational price. 

\subsection{Comparison with the Lasso and random forests}
We compare in this last subsection the performance of the standard and accelerated boosting algorithms with that of the Lasso and random forests, respectively implemented with the \verb+R+ packages \verb+glmnet+ and \verb+randomForest+. As above, the number of components $T^{\star}$ of the boosting predictors are selected by minimizing \eqref{eq:sel_iter_test}. The shrinkage parameter of the Lasso (parameter \verb+lambda+ in \verb+glmnet+) and the number of variables randomly sampled as candidates at each split for the trees of the random forests (parameter \verb+mtry+ in \verb+randomForest+) are selected by minimizing the mean squared error (regression) and the misclassification error (classification) computed on the validation set. The \verb+R+-package \verb+caret+ was used to conduct these minimization problems. The prediction performance of each predictor $F$ were assessed on the test set by the mean squared error
$\frac{1}{\sharp \mathscr D_{{\rm test}}}\sum_{i\in\mathscr D_{{\rm test}}}(Y_i-F(X_i))^2$
for regression problems, and $(i)$ the misclassification error $\frac{1}{\sharp \mathscr D_{{\rm test}}}\sum_{i\in\mathscr D_{{\rm test}}}\mathds 1_{F(X_i)\neq Y_i}$ and $(ii)$ the area under ROC curve (AUC) for classification problems (computed on the test set).

Table \ref{tab:quad_risk_10mod} shows the test errors for the regression problems, while Tables \ref{tab:missclass_10mod} and \ref{tab:auc_10mod} display misclassification errors and AUC for classification tasks. All results are averaged over 100 replications for simulated examples and over 20 permutations of the sample for real-life data set. 

As might be expected, the results depend on the data sets, with an advantage to boosting algorithms, which are often the first and perform uniformly well. Besides, even if there is no clear winner between traditional boosting and \verb+AGB+, we still find that \verb+AGB+ is weakly sensitive to the choice of $\nu$ and leads to more parsimonious models ($T^{\star}$ in the tables) for both regression and classification problems, and independently of the data set. They are the take-home messages of our paper.

\begin{table}[!h]
  \centering
{\scriptsize
  \begin{tabular}{|cc||c|c|c|c|c||c|c|c|c|c||c||c|}
\hline
  &  &\multicolumn{5}{c||}{{\bf GB}}&\multicolumn{5}{c||}{{\bf AGB}}& {\bf Lasso} & {\bf RF} \\
& $\nu$ & 1e-05 & 0.001& 0.01 & 0.1 & 0.5 & 1e-05 & 0.001& 0.01 & 0.1 & 0.5 & & \\
\hline
\hline
Model 1 (u)& m. & 1.011 &   {\bf 0.923} &  0.926 & 0.927 & 0.930 &    0.924  &   0.927  &  0.926 &  0.929  &  {\bf 0.920} & 1.021& 0.922 \\
   & sd. & 0.082  &  0.075 &  0.076 & 0.076 & 0.079  &   0.076  &   0.077 &   0.074 &  0.074 &  0.081 & 0.084 & 0.078 \\
& $T^\star$ & 10 000  &   7 924  &   981 &    99 &    11 &     2 178  &     247  &     73  &    18  &     7 & & \\
\hline
Model 2 (u)& m. & 1.883 &   0.642 &  {\bf 0.621} & 0.621 & 0.650  &   0.632  &   {\bf 0.621} &   0.621 &  0.638  & 0.794 & 0.677 & 0.756 \\
   & sd. & 0.202 &   0.073 &  0.075 & 0.074 & 0.079  &   0.069  &   0.072 &   0.072 &  0.073 &  0.090 & 0.077 & 0.086 \\
& $T^\star$ & 10 000  &   9 989  &  2 206 &   214  &   26   &   2 488  &     288  &     91 &     26  &    14 & & \\
\hline
Model 3 (u)& m. & 2.983  &  0.318 &  {\bf 0.037} & 0.040 & 0.119  &   0.308 &    0.080  &  {\bf 0.078} &  0.125 &  0.337 & 0.948 & 0.587 \\
   & sd. &  0.221  &  0.039 &  0.007 & 0.008 & 0.015  &   0.042  &   0.019  &  0.017 &  0.020 &  0.060 & 0.067 & 0.068 \\
& $T^\star$ & 10 000  &  10000  &  7 936  &  956  &   97  &    2 500   &    627  &    187   &   49  &    29 & & \\
\hline
\hline
Model 1 (c)& m. & 8.316 &   4.483 &  {\bf 4.047} & 4.051 & 4.220  &   4.529  &   4.141  &  {\bf 4.133} &  4.252 &  5.354 & 8.549 & 4.163 \\
   & sd. & 1.143  &  0.668 &  0.557 & 0.559 & 0.573  &   0.669 &    0.564 &   0.566 &  0.575 &  0.694 & 1.154 & 0.623\\
& $T^\star$ & 10 000  &  10 000  &  3 413  &  330  &   47  &    2 500  &     387  &    120   &   32  &    12 & & \\
\hline
Model 2 (c)& m. & 6.558  &  2.424  & {\bf 1.936}  & 1.938 & 2.093  &   2.442 &    2.083  &  {\bf 2.057} &  2.145 &  2.777 & 4.988 & 2.082  \\
   & sd. & 1.958  & 1.120  & 1.093 & 1.095 & 1.118  &   1.117  &   1.087  &  1.087 &  1.062 &  1.103 & 1.580 & 0.824 \\
& $T^\star$ & 9 900  &  10 000  &  4 632 &   458  &   70  &    2 499   &    411 &     132   &   35   &   16 & & \\
\hline
Model 3 (c) & m. & 37.034 &   6.323 &  {\bf 4.454} & 4.480 & 5.879  &   6.382  &   5.274  &  {\bf 5.163} &  5.781 &  8.187 & 23.898 & 6.198 \\
   & sd. &  8.617 &   3.883 &  3.703 & 3.708 & 3.948  &   3.936  &   3.824 &   3.761 &  3.827 &  4.020 & 5.746 & 3.421 \\
& $T^{\star}$ & 10 000  &  10 000  &  4 296 &   415  &   54   &   2 491   &    361 &     113  &      31  &    23 & & \\
\hline
\hline
Crimes& m. & 0.049 &   0.019 &  0.019 & 0.019 & 0.021  &   0.021  &   0.021  &  0.021 &  0.021  & 0.024 & {\bf 0.019} & {\bf 0.019} \\
   & sd. & 0.004  &  0.001 &  0.001 & 0.002 & 0.002  &   0.002  &   0.001  &  0.001 &  0.002 &  0.002 & 0.001&  0.001 \\
 & $T^\star$ & 10 000  &   9 960 &   2 172  &  214  &   86  &    2 240   &    296   &    91  &    26  &    16 & & \\
\hline
Wine& m. & 0.632 &   0.417 &  {\bf 0.412} & 0.412 & 0.419  &   0.421  &   {\bf 0.421}  &  0.421 &  0.424  & 0.459 & 0.426 & 0.365 \\
   & sd. & 0.044 &   0.032 &  0.032 & 0.032 & 0.032  &   0.034  &   0.033  &  0.032 &  0.032 &  0.034 & 0.001 & 0.001 \\
 & $T^\star$ & 10 000  &   9 999 &   3727  &  366  &   79  &    2 433  &     393   &   154  &    36  &    11 & &  \\
\hline
\hline
  \end{tabular}
  \medskip
  \caption{Mean (m.) and standard deviation (sd.) of the mean squared test error for the regression problems. Also shown for the boosting algorithms is the mean over all replications of the optimal number of components ($T^\star$) . Results are averaged over 100 independent replications for simulated examples and over 20 independent permutations of the sample for real-life data sets. For each data set, the two best performances are in bold.}
  \label{tab:quad_risk_10mod}
}
\end{table}

\begin{table}[!h]
  \centering
{\scriptsize
  \begin{tabular}{|cc||c|c|c|c|c||c|c|c|c|c||c||c|}
\hline
  &  &\multicolumn{5}{c||}{{\bf GB}}&\multicolumn{5}{c||}{{\bf AGB}}& {\bf Lasso} & {\bf RF} \\
& $\nu$ & 1e-05 & 0.001& 0.01 & 0.1 & 0.5 &  1e-05 & 0.001 & 0.01 & 0.1 & 0.5 & & \\
\hline
\hline
Model 4 (u)& m. & 0.416  &  0.229 &  0.098 & 0.086 & {\bf 0.085}   &  0.248 &    {\bf 0.085}  &  0.088 &  0.108 &  0.217 & 0.419 & 0.206 \\
   & sd. &  0.020 &   0.023 &  0.018 & 0.015 & 0.016  &   0.023  &   0.016  &  0.016 &  0.017 &  0.036 & 0.021 & 0.025 \\
& $T^\star$ & 9 900 &   10 000  &  9 998 &  2 619  &  452  &    2 500  &    1 404   &   421  &    97  &    22 & & \\
\hline
Model 5 (u) & m. & 0.353  &  0.144 &  {\bf 0.141} & 0.141 & 0.142  &   0.145  &   0.142 &   {\bf 0.141} &  0.144 &  0.155 & {\bf 0.138} & 0.151 \\
   & sd. & 0.024  &  0.016 &  0.017 & 0.016 & 0.018  &   0.017  &   0.018  &  0.017  & 0.016  & 0.021 & 0.018 & 0.019 \\
& $T^\star$ & 10 000  &  10 000  &  2 465  &  240  &   41   &   2 500  &     387  &    121  &    34   &   12 & & \\
 \hline
\hline
Model 4 (c)& m. & 0.451  &  0.171  & 0.086 & 0.081 & {\bf 0.079}  &   0.185  &   {\bf 0.080}  &  0.081  & 0.095 &  0.183 & 0.453 & 0.134\\
   & sd. & 0.027 &  0.020 &  0.015 & 0.014 & 0.014  &   0.022  &   0.014 &   0.015 &  0.015 &   0.03 & 0.025 & 0.018\\
& $T^\star$ & 10 000  &  10 000  &  9 996 &  1 781  &  319   &   2 500   &   1 156  &    358  &    88   &   23 & & \\
\hline
Model 5 (c)& m. & 0.423 &   0.119 &  {\bf 0.114} & 0.114 & 0.115  &   0.123  &   {\bf 0.114}  &  0.116  & 0.118 &  0.132 & 0.118 & 0.116\\
   & sd. &  0.037  &  0.016 &  0.015 & 0.016 & 0.016  &   0.018  &   0.016  &  0.016  & 0.016 &   0.020 & 0.016 & 0.018 \\
& $T^\star$& 10 000  &  10 000  &  3 694 &   354  &   65  &    2 500   &    493  &    151   &   40  &    14 & & \\
\hline
\hline
Adult & m. & 0.249  &  0.150 &  0.141 & {\bf 0.138} & 0.138  &   0.151  &   {\bf 0.140}  &  0.140  & 0.143 &  0.152 & 0.155 & 0.186 \\
   & sd. & 0.004  &  0.004 &  0.004 & 0.004 & 0.004  &   0.004  &   0.005 &   0.005 &  0.004 &  0.004 & 0.004 & 0.005 \\
& $T^\star$ & 10 000  &  10 000  &  9 966 &  6 714  & 1 635  &    2 500   &   1 853  &    610  &   143  &     24 & & \\
\hline
Advert& m. & 0.165  &  0.062 &  0.043 & 0.043 & 0.043  &   0.063  &   0.043 &   0.043 &  0.044 &  0.054 & {\bf 0.032} & {\bf 0.031} \\
   & sd. & 0.014 &   0.009 &  0.012 & 0.013 & 0.012  &   0.008  &   0.013  &  0.013 &  0.011 &  0.011 & 0.007 & 0.009 \\
& $T^\star$ & 10 000  &   9 999  &  4 716  &  471  &   87  &    2 500   &    568   &   181  &    50   &   18 & & \\
 \hline
Spam& m. & 0.396  &  0.071  & {\bf 0.061} & 0.061 & 0.065  &   0.077  &   0.064  &  0.065 &  0.068 &  0.086 & 0.095 & {\bf 0.057} \\
   & sd. & 0.013  &  0.009 &  0.008 & 0.007 & 0.007  &   0.009  &   0.009  &  0.007 &  0.007 &  0.011 & 0.072 & 0.007 \\
& $T^\star$ & 10 000  &  10 000  &  3 880  &  426  &   84  &    2 500   &    479  &    150  &    40   &   16 & & \\
\hline
\hline
  \end{tabular}
  \medskip
  \caption{Mean (m.) and standard deviation (sd.) of the misclassification test errors for the classification problems. Also shown for the boosting algorithms is the mean over all replications of the optimal number of components ($T^\star$) . Results are averaged over 100 independent replications for simulated examples and over 20 independent permutations of the sample for real-life data sets. For each data set, the two best performances are in bold.}
\label{tab:missclass_10mod}
}
\end{table}

\begin{table}[!h]
  \centering
{\scriptsize
  \begin{tabular}{|cc||c|c|c|c|c||c|c|c|c|c||c||c|}
\hline
  &  &\multicolumn{5}{c||}{{\bf GB}}&\multicolumn{5}{c||}{{\bf AGB}}& {\bf Lasso} & {\bf RF} \\
&$\nu$ & 1e-05 & 0.001& 0.01 & 0.1 & 0.5 &  1e-05 & 0.001 & 0.01 & 0.1 & 0.5 & & \\
\hline
\hline
Model 4 (u)& m. & 0.590  &  0.885 &  0.971 & 0.976  & {\bf 0.977}  &   0.869  &   {\bf 0.977} &   0.975 &  0.964 &  0.862 &0.515  & 0.891 \\
   & sd. & 0.037 &   0.021 &  0.008 & 0.007 & 0.007  &   0.023  &   0.007 &   0.007 &  0.010  & 0.040 & 0.018 & 0.021 \\
& $T^\star$ & 9 900  &  10 000  &  9 998 &  2 619  &  452  &    2 500  &    1404  &    421  &    97   &   22 & & \\
\hline
Model 5 (u)& m. & 0.772 &   0.935 &  0.936 & 0.936 & 0.934  &   0.933  &   {\bf  0.937}  &  0.936 &  0.935 &  0.922 & {\bf 0.940} & 0.922 \\
   & sd. & 0.059  &  0.013 &  0.012 & 0.012 & 0.013  &   0.013 &    0.013  &  0.012 &  0.012 &  0.015 & 0.011 & 0.016 \\
& $T^\star$ & 10 000  &  10 000  &  2 465  &  240  &   41   &   2 500    &   387   &   121   &   34  &    12 & & \\
\hline
\hline
Model 4 (c)& m. & 0.621 &   0.927 &  0.978 & {\bf 0.981} & 0.981  &   0.916  &   {\bf 0.981}  &  0.981 &  0.972 &  0.898 & 0.516 & 0.945\\
   & sd. & 0.043  &  0.014 &  0.006 & 0.005 & 0.005  &   0.016  &   0.005  &  0.005 &  0.008 &  0.030 & 0.019 & 0.012 \\
& $T^\star$& 10 000  &  10 000  &  9 996 &  1 781 &   319  &    2 500   &   1 156  &    358   &   88  &    23 & & \\
\hline
Model 5 (c)& m. & 0.753  &  0.960 & 0.962 & {\bf 0.963} & 0.961   &  0.957  &   {\bf 0.962} &   0.962 &  0.960 &  0.947 & 0.960 & 0.955 \\
   & sd. & 0.059  &  0.009 &  0.008 & 0.008 & 0.008  &   0.009 &    0.008  &  0.008  & 0.008 &  0.011 & 0.007& 0.011\\
& $T^\star$& 10 000  &  10 000  &  3 694  &  354  &   65  &    2 500   &    493  &    151  &    40  &    14 & & \\
\hline
\hline
Adult & m. & 0.758  &  0.905 &  0.915 & {\bf 0.920} & 0.920 &    0.902  &   {\bf 0.918} &   0.917 &  0.913 &  0.901 & 0.902 & 0.858 \\
   & sd. & 0.005 &   0.004 &  0.004 & 0.004 & 0.003  &  0.004  &   0.004  &  0.004 &  0.003 &  0.004 & 0.004 & 0.008 \\
& $T^\star$ & 10 000  &  10 000  &  9 966 &  6 714  & 1 635  &    2 500   &   1 853  &    610  &   143  &    24 & & \\
\hline
Advert & m. & 0.815 &   0.962 &  0.974 & 0.973 & 0.973  &   0.956  &   0.973  &  {\bf 0.975} &  0.971 &  0.950 & 0.973 & {\bf 0.983} \\
 & sd. & 0.059  &  0.014 &  0.011 & 0.012 & 0.013  &   0.015  &   0.014  &  0.011 &  0.015 &  0.022 & 0.008 & 0.008 \\
& $T^\star$ & 10 000  &   9999  &  4716  &  471  &   87  &    2500  &     568   &   181  &    50  &    18 & & \\
\hline
Spam& m. & 0.854  &  0.975 &  {\bf 0.980} & 0.980 & 0.979  &   0.973  &   0.978  &  0.978  & 0.977 &  0.966 & 0.970 & {\bf 0.979} \\
   & sd. & 0.028 &   0.003  & 0.003 & 0.003 & 0.003  &   0.004  &   0.003  &  0.003 &  0.003 &  0.005 & 0.004 & 0.003 \\
& $T^\star$ & 10 000  &  10 000  &  3 880  &  426  &   84   &   2 500   &    479   &   150   &   40  &    16 & & \\
\hline
\hline
  \end{tabular}
  \medskip 
  \caption{Mean (m.) and standard deviation (sd.) of AUC for the classification problems. Also shown for the boosting algorithms is the mean over all replications of the optimal number of components ($T^\star$) . Results are averaged over 100 independent replications for simulated examples and over 20 independent permutations of the sample for real-life data sets. For each data set, the two best performances are in bold.}
  \label{tab:auc_10mod}
}
\end{table}

\newpage

\bibliography{biblio-agb}

\end{document}